\documentclass[letterpaper, 10 pt, conference]{ieeeconf}

\IEEEoverridecommandlockouts                              

\usepackage{graphicx} 

\title{Multiple criteria decision-making for lane-change model
}

\author{Ao Li$^{1}$ Liting Sun$^{2}$ Wei Zhan$^{2}$ Masayoshi Tomizuka$^{2}$   
\thanks{$^{1}$Ao Li is visiting student researcher of the Department
of Mechanical Engineering, University of California, Berkeley, CA
94720 USA.
        {\tt\small aoli@berkeley.edu}}%
\thanks{$^{2}$L. Sun, W. Zhan and M. Tomizuka are with the Department
of Mechanical Engineering, University of California, Berkeley, CA
94720 USA.
        {\tt\small {litingsun, wzhan, tomizuka}@berkeley.edu}}%
}

\begin{document}

\maketitle
\thispagestyle{empty}
\pagestyle{empty}

\begin{abstract}
Simulation has long been an essential part of testing autonomous driving systems, but only recently has simulation been useful for building and training self-driving vehicles. Vehicle behavioural models are necessary to simulate the interactions between robot cars. This paper proposed a new method to formalize the lane-changing model in urban driving scenarios. We define human incentives from different perspectives, speed incentive, route change incentive, comfort incentive and courtesy incentive etc. We applied a decision-theoretical tool, called Multi-Criteria Decision Making (MCDM) to take these incentive policies into account. The strategy of combination is according to different driving style which varies for each driving. Thus a lane-changing decision selection algorithm is proposed. Not only our method allow to vary the motivation of lane-changing from purely egoistic desire to a more courtesy concern, but also they can mimic drivers' state, inattentive or concentrate, which influences their driving Behaviour. We define some cost functions and calibrate the parameters with different scenarios traffic data. Distinguishing driving styles are used to aggregate decision-makers' assessments about various criteria weightings to obtain the action drivers desire most. Our result demonstrates the proposed method can produce varied lane-changing behaviour. Unlike other lane-changing models based on artificial intelligence methods, our model has a more flexible controllability.

\end{abstract}


\section{INTRODUCTION}
Lane-changing models are a vital component in autonomous driving simulation tools, which are extensively used and playing an increasingly important role in behaviour studies. With the rise of autonomous driving research, the demand for establishing car behaviour model in simulators \cite{dosovitskiy2017carla} \cite{naumann2018coincar} in this area has increased. Furthermore, microscopic traffic models can be used together in simulator to create virtual scenarios in which the lane-changing model is an essential component for replicating real-world individuals behaviours. As car-following model can simplify the kinematic model and mimic the longitudinal motion along the lane in simulator, lane-changing model are supposed to produce variant intention towards to lane-changing decisions.

Lane-changing model has been developed since 1960s, they could be classified into several types: rule-based, discrete choice model, Artificial Intelligence model and incentive-based model \cite{rahman2013review}. MOBIL \cite{kesting2007general} is a classical incentive-based model which takes the safety and acceleration into account, this attribute makes it only suitable for highway scenarios. However, lane-changing behavior in urban scenarios is much more complex to evaluate due to the complicated driver style and environmental impacts, such as the driver's age, sex, career, character and some factors of the roadway, which have not been adequately considered in existing models. We proposed to evaluate the lane-changing probability by combining the driving style and incentives

In this paper, a more comprehensive and macroscopic lane-changing model is proposed for simulating urban driving behaviours. Some possible lane-changing incentives, comfort, route, speed and courtesy, are discussed and formulated mathematically by evaluation function. A multiple-criteria decision making method is used to integrate these factors together to improve the accuracy and reliability. Different driving styles are integrated into our model. These distinguishing styles could be used to tune weights in the evaluation of lane-changing desire. All parameters are calibrated in our simulator by traffic data via different scenarios, high way and urban. And the tuning of parameters can produce abundant and unexpected lane-changing behaviours, which demonstrates the flexible and versatile feature of our model.

\section{Related work}
In the past, lane-changing model is well studied in different perspectives. Ahmed presented in 1999 a utility-based framework which divides lane-changing maneuvers into mandatory (MLC) and discretionary (DLC) and adopts a decision three to model a lane-changing maneuver\cite{kita1999merging}. 
Another lane-changing  model that focused on interactions between a merging vehicle and through vehicles in an on-ramp location is Kita’s model\cite{kita1999merging}. Some of the recent emerged lane-changing models use artificial intelligence methods to make lane-changing decision, such as randomized forest\cite{jiang2014study} or layered perceptron \cite{ren2019new}. Deep learning-based decision making methods are widely studied in recent years, the method can produce abundant and reasonable result. But drivers' decision is highly related to their driving style, we can hardly modify their models' parameters to mimic the specific driving styles because of the uncontrollability of DL model.

Incentives of lane-changing are complex and have been studied for a long time. \cite{kesting2007general} formulated a lane change model combined with incentives to safety and minimize brake cost. \cite{schakel2012integrated} proposed a incentive-based model which combined three main aspects, follow a route, to gain speed, and to keep right, for determining a lane change desire.

Driving style is also studied in different perspective. It can affect any action during the cruise and its classification are also various. The extent of aggressiveness can be divided as cautious, stable or adventitious. In terms of the interaction of others, they can be divided as egoism vs altruism. Altruists are polite drivers who consider more others' interests. Additionally, drivers could be inattentive due to distraction or fatigue \cite{meiring2015review}.

\subsection{Car-following models review}
As lane-changing model is in the level of decision-making, a dynamic kinematic model should be integrated to describe the longitudinal motion of vehicles. Thus car-following models will be discussed in this section. In the past a variety of car-following models were established, They are derived from different perspectives and are suitable in different traffic situation. Next, we mainly introduce the two families of the car-following model.

\subsubsection{OVM}
Optimal velocity model is proposed by \cite{bando1998analysis}, this model uses a optimal velocity determined by the inter-vehicle distance to present drivers' velocity anticipation. \\
This Optimal Velocity Model describes the following features,
\begin{itemize}
    \item A car will keep the maximum speed with enough the distance to the next car.
    \item A car tries to run with optimal velocity determined by the distance to the next car.
\end{itemize}

These features make it suitable for adventitious drivers who dare to keep an optimal velocity to reduce their driving time. But this model encounters the problems of too high acceleration and unrealistic deceleration. In order to solve the problems, Generalized forces model \cite{helbing1998generalized}(GFM) was proposed. It inherits the optimal velocity concept from OVM and is also derived from the pedestrian Social forced model. It proposed the virutal social forces concept into the traffic network. Each car is influenced by repulsive and attractive forces. Full Velocity Difference model (FVDM) \cite{jiang2001full} is an enhance GFM, it takes both positive and negative velocity differences with the precedent car into account. The main advantage of FVDM is eliminating unrealistically high acceleration and predicts a correct delay time of car motion and kinematic wave speed at jam density. However, the velocity difference is not enough to avoid an accident under an urgent case.


\subsubsection{IDM}
The intelligent driver model (IDM) is a time-continuous car-following model for the simulation of freeway and urban traffic. It was developed by \cite{treiber2000congested} and is a avoidance of collision model cause it takes safe time headway into count and delimiting unrealistic acceleration and deceleration. The non-avoidance feature makes IDM a conservative driving behaviour model. \cite{kesting2010enhanced} proposed an enhanced IDM to eliminate some unrealistic behaviour in lane-changing in congested traffic.

Different car-following models could produce different resulting accelerations and velocity in similar situation due to their own characteristics. Thus they can be used to fit different driving styles as an underlying model with our lane-changing model.

\section{Multiple incentives decision analysis}
Incentives could be view as the basic elements in decision-making in sociology, economics and engineering etc. People have different criterias and responses for a same incentive.

In classic MOBIL\cite{kesting2007general} model, the acceleration incentive criterion measures the attractiveness of a given lane based on its utility, and the safety criterion measures the risk associated with lane changing (i.e., acceleration). They have a hierarchy as acceleration incentive is firstly considered.

In practice, we have more macroscopic and long-term incentives to drive us make a lane-changing decision. In our method, safety criteria is regarded as a low-level condition we should pay attention while we are executing lane-changing behaviour. Some other more high-level motivations, such as velocity, acceleration gain and the pursuit of comfort are developed and combined by MCDM.

Most of people prefer to choose their custom route in daily life. But sometimes they tend to change to a a less congested route to avoid traffic. The travel time of each route are accessible via the navigation system in our car. depends not only on traffic guidance system. We may sometimes choose to not follow the recommended routing path in navigation app for an avoidance of a traffic jam. Someone tired may prefer to keep the original route because a lane-changing progress costs drivers' too much attention, they will be forced to notice carefully the feasible gap and vehicles' speed and synchronized with the velocity of target lane. Even they are involved into a game theory with other driver because they have to estimate other drivers intention(yield or not yield) while they are try to merge. Thus drives who are tired or care little about the time cost, they often regard the driving as a relaxation and hardly like to be nervous.

We assume that the drivers in transportation system are bounded rational , they make decision according they driving styles. And the traffic information they received is incomplete. Besides the navigation system can provide route travel time, people can only use the surrounding situation in their sight to estimate the whole community traffic. 

Mandatory lane-changing decision may disable the incentive policy. In our studies we eliminate mandatory lane-changing. Everyone take lane-changing action as voluntary desire. In our method, we can generate agent with different characters by varying the parameters of driving style so that each agent can use their own values to judge whether it is worthwhile to change lanes.

\section{Policies}

In community-size driving scenarios, each lane may led to different road section in the intersection. If the current route cannot  be followed any more on the target lane. Thus people change to the lane next to lane may bring them a result that they cannot get back timely. The adjacent and parallel lanes which led consequently to the same section can be viewed as symmetric lanes. The lane-changing behaviour between symmetric lanes can be simplified as a pure velocity gain incentives as MOBIL does. But change to a asymmetric lane may involved more incentives such as change of routing path or overtaking some precedent cars.

\begin{figure}[t]
\begin{center}
  \includegraphics[width=0.7\linewidth]{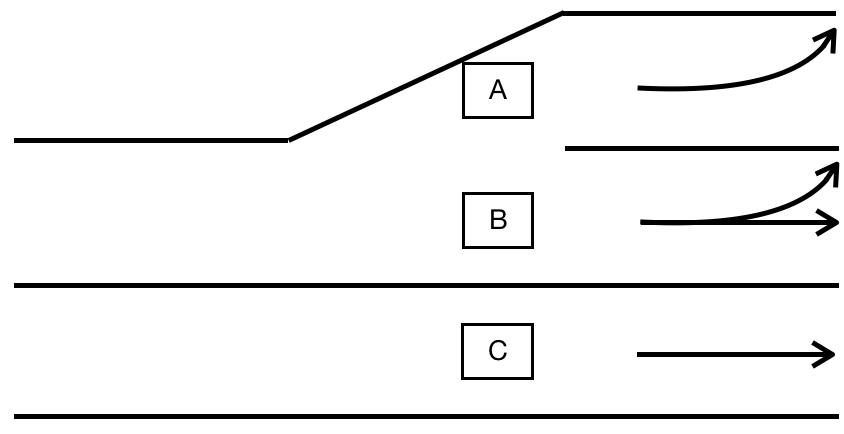}
\end{center}
  \caption{Illustration of symmetric vs asymmetric lane. Lane B leads to both left and straight forward, thus B are symmetric for Lane A and Lane C. But A and C are asymmetric for each other, additionally for B, they are both asymmetric.}
\label{fig:long}
\label{fig:onecol}
\end{figure}

In the figure, the lane b and c have the same following section, thus they are symmetric. The lane change-behaviour between these two lanes only involves the acceleration or velocity gain. Once people drive into lane a might bring them into a new routing path.

\subsection{Probability of back}
The difficulty of changing lanes is positively related to the traffic density of the target lane. If the target lane is almost vacuum, the degree of difficulty tend to be 0. In this case, drivers more likely tend to change lane to overtake the precedent car, they definitely could change back to the original route before they reach the stop line or off-ramp. Plus, drivers have more chance to change back if the remaining distance is long enough. In other words, people take a risk of route change if they change lane to an asymmetric lane unless they determine to change their original route to avoid traffic congestion. Thus a routing path change behavior may lead to a lane change action, vice versa.
We propose a parameter $P_b$ that indicates the probability if the drivers can change back. If they change back to their original lane, the double changing behaviour can be viewed as a overtaking. If not, the driver may be voluntary or forced to take a new routing path.

This probability is related to average gap distance and vehicles velocity in the target lane. The average of time headway : $$ T_h = \frac{\bar{d}}{\bar{v}} $$ where the $ \bar{d} $ is the mean headway in this lane. And more long length the parallel roads have, the more chance drivers will have to back. Then we can define an equation that indicate the probability of back to the original lane. $$P = 1 - e^{-\alpha*T_{h}*S} $$ where $T_h$ is the average of time headway and $S$ is the remaining distance.
\subsection{Route change incentive}
People normally has a desire to keep in shortest routing path. But sometimes while a driving in a hurry meets traffic jam in ego lane, he might decides to change other routing path to avoid this.

Traffic condition is always complicated and uncertain, and the travelers cannot accurately perceive the current or future traffic conditions. Thus, drivers make their route choice decisions under risk and uncertainty. As all vehicles equipped a navigation system, drivers have a rough estimation of all possible paths and choose their desired path instead of the shortest path. 

In other aspect, as we discussed in the previous section, people change lane may result finally a change of routing path cause they cannot change back timely. In order to resist this current risk, the potential path should be put in drivers' mind. For example, if people miss an exit on the highway, they may have to get some kilometers more to reach the destination. In this case, we should give a huge penalty.

Routing cost can be qualified by the length of route, the number of turns and number of traffic lights etc. Here we use the travel time to evaluate the cost of route.

To calculate the average travel time, we use the function provided by \cite{guessous2014estimating}: \\
$$ T_l = T_0(1+k_1(\frac{q_l}{q_{max})^k_2} ) $$ \\
where $l$ is the level of service which describes the traffic density of a road. $k_1$ and $k_2$ are dimensionless parameters given by the original paper.

\subsection{Speed incentive}
This incentive is most largely discussed in other lane-changing model\cite{kesting2007general} \cite{schakel2012integrated}. In microscopic scenario, drivers use surrounding cars to estimate their potential gain if they execute a lane-changing. A car-following model is integrated to calculate the longitudinal acceleration and expected velocity. The desired velocity which reflects the driver's aggressiveness is took into account by the car-following equations. The current speed to a large extent is decide by the head vehicle. It can be calculated by IDM equation:
$$ V = V_{IDM} = V + a (1 - (\frac{v_\alpha}{v_0})^\beta - (\frac{s^*}{s_\alpha})^2) $$
A driver change lane for a speed gain tend to be adventitious and aggressive, FVDM is more suitable to estimate his expected speed 
$$ V = V_{OVM} = V_1 + V_2tanh(C_1(s - l_c) ~ C_2) $$

Full Velocity Difference Model should integrated as underlying model instead of IDM in the agents which are aggressive because they act more adventitiously with a pursuit of optimal velocity. 

\subsection{Courtesy incentive}
Some cars with politeness attributes may give way to the following blocked cars by changing lane. Or in some ramp-on situations, they change lane to give more space for others vehicles facing a mandatory lane-changing. Altruism takes more attention at others' benefits. The closer they are to the merging intersection, the greater the probability that they will give way.

Two cases are specified: 

1. The ego velocity is slower than the average velocity of the current lane. Altruism driver may give way to the following car to make them not to be blocked.

After the end of the lane change, the current followers and new followers will both a new speed $\widetilde{V_o} $ and $\widetilde{V_n} $.
Where $\widetilde{V_o} $ is the new velocity of old successor, and $\widetilde{V_n}$ is the new velocity of new successor.These two velocities are estimated by a car-following model. \\
$$ A = \frac{| v_s - {v_{s\_desired}} | - |\widetilde{v_s}-{v_{s\_desired}}|}{ |v_s-{v_{s\_desired}}|} 
+$$ $$ \frac{| v_s - {v_{s\_desired}} | - |\widetilde{v_s}-{v_{s\_desired}}|}{ |v_s-{v_{s\_desired}}|}$$

2. More closer to a on-ramp or off-ramp scenarios, more possible drivers who change to the inner lane unless their route are bound to the current lane. This scenario can also be also considered as the first one. While other vehicles are trying to merge into the lane of ego-vehicle, we can regard these vehicles as our related vehicles, our courtesy incentive is committed to make related car have the velocity closer to desired. Thus the courtesy gain can be generalized as:
$$ A = \sum{\frac{| v_r - {v_{r\_desired}} | - |\widetilde{v_r}-{v_{r\_desired}}|}{ |v_r-{v_{r\_desired}}|} }$$
Where the $v_r$ is the velocity of each vehicle related to ego-vehicle. This cost can present the extent not to hinder others vehicle in the road.
\subsection{Comfort incentive}
People would have anxiety about traffic accidents if they change lane in a congested traffic condition. As to a driver, lane change behaviour cost him so much attention which makes him nervous. In traditional motion planning algorithm, people use the jerk and yaw rate to calculate the comfort. These parameters can only demonstrate the passengers' comfort instead of drivers'. 

Drivers' comfort is much more related to the average time headway of the target lane. If the time headway is long enough, drivers don't need to pay attention at the accepted gap distance. Conversely, under a traffic whose average headway is short, people have to synchronize their motion with the target lane. Once the headway is less than a threshold, drivers have to attempt to cut in the target lane by adjusting their planned path again and again and do cooperative planning with the successor driver. This progress can sharply augment the drivers' tension.
As less average time headway is, more the times cooperate other vehicles are. And the difficulty of manipulation in the lane change will increase with the decrease of time headway. We suppose comfort cost of lane-changing: 
$$ J = - K * T_{h}^{-\beta} $$where $T_h$ is the average of time headway of the target lane. 
During the combination by multiple criteria decision making, the parameter K can be retrieved out and integrated into the weighting coefficients.

Comfort cost is aimed to mimic the drivers who has an inattentive driving style or someone tired. They don't care too much about the time cost and would like to make driving as relax as possible.
\subsection{safety criteria}
Safety criteria refers to the gap acceptance for a lane-changing behaviour. The critical gap distance can varies by several parameters \cite{toledo2003modeling} in different traffic condition. To simplify, we regard a gap distance is acceptable once if the time-to-collision between the new follower and ego car is greater than a threshold $T_{th}$.

In traditional probabilistic lane-changing model. The lane-changing behaviour is model as a decision tree and gap acceptance is always in the lowest layer. This is also similar in MOBIL, it firstly considered incentive is also acceleration gain and then the safety criteria.

We also regard safe criteria as a temporary condition instead of an incentive. Once drivers make a lane-changing decision, he should notice the gap on the target lane. If the gap distance is not acceptable for a totally safe lane-changing, they should synchronize with the vehicles of target lane or keep waiting.


\subsection{Combine Criteria with driving style}
MCDM is one of the most widely used decision methodologies in engineering, technology, science and management and business. In MCDM domain, several families of combination strategy are proposed. We choose linear combination strategy due to a complex formulation, such as exponential relation\cite{gonzalez2002navigation}, will sharply increase the difficulty of parameters' calibration.

Different drivers have different evaluation standard and our criteria policies so highly related to driving style that these weighs are the interface to mimic different kind of driving style. For example, a greater weight of routing incentive indicates the driver has preference of time saving. And the agent which pay more attention at speed gain appears more aggressive in urban traffic environment.

Driving style concerns the way a driver chooses to drive, and depends on physical and emotional conditions of the driver while
driving. Driving style and driving skill can be assumed as the fundamental factor of different driving behaviours \cite{elander1993behavioral}. They might be able to illustrate the behaviour differences of one driver from another in the same vehicle in the same situations. It both influences the car-following behaviour \cite{ishibashi2007indices} and particularly for the lane-changing behaviour.

The classification of driving style and its impacts are studied in \cite{meiring2015review} \cite{martinez2017driving}. Conservative behaviour could be categorized as safe driving style, they hardly pursuit a irregular speed or overtake other vehicles. Aggressive driving style seems to have the opposite characteristics. We can give a larger desired velocity and heavier weights of speed incentive into these agents. Inattentive driving style indicates regular inattention to driving actions and necessary observations to complete the driving task. Driver fatigue and driver distraction could result their inattention \cite{dong2010driver}. In practice, these people prefer to follow their custom or original route and are not willing to make then nervous. Thus we can augment the comfort incentive weights to emulate inattentive style. As to the interaction between vehicles, drivers' characters can be regarded egoism, altruism between the two. The coefficient of courtesy can present how much the driver cares about the others' benefits. As we discussed previously, people change to asymmetric lane with a possibility that they might change their route. Thus for each lane-changing behaviour, we have three possible results:
\begin{itemize}
    \item Keep on the current lane $C_o$.
    \item Change lane and get back timely $C_o$, which can be viewed as overtaking.
    \item Change both lane and routing path $C_{cr}$.
\end{itemize}
In the case that the target lane is symmetric to the current, routing cost are the same because they have no sign of changing the original route (symmetric lanes lead to the same section).

\begin{table}[h]
\begin{center}
  \caption{Different costs in all possible cases \label{tab:reg}}
    \begin{tabular}{|l|ccc|}
    \hline
    Cases  & $C_o$  & $C_o$ & $C_{cr}$    \\
    \hline 
    Possibility     & $-$   & $P_{back}$   & $ 1-P_{back}$ \\
    Routing         & $ T(S, V) $  & $ T(S, \widetilde{V})$ & $ T(\widetilde{S}, \widetilde{V}) $ \\ 
    Speed           & $ V $     & $ \widetilde{V} $  & $ \widetilde{V} $   \\ 
    Courtesy        & $ \sum{V_r} $  & $ \sum{\widetilde{V_r}} $ & $ \sum{\widetilde{V_r}} $   \\ 
    Comfort         & $0$ & $ \widetilde{J}+J $ &  $ \widetilde{J}$ \\ 
    \hline
    \multicolumn{4}{l}{\scriptsize Symbols with a tilde means the resulting cost after a lane-changing.}
    \end{tabular}
\end{center}
\end{table}

The expectation of the lane-changing gain is:
$$ C = p*C_{cl} + (1-p)*C_{cr} - C_o $$ 
As these cost criteria has different scale. We use the earning rate to describe the attractiveness of lane-changing.
$$ Y = \frac{C}{C_o} $$
We can combine these earning yields linearly: $$G = \sum{\mu_iY_i}$$
The total earning yield determines the desired degree of lane-changing. Should note that the weights of each incentive reflect the driving style of a driver. They are not bounded to a fixed set but we can adjust them to achieve different preference of a driver. Once combined earning yields $G > threshold$, the robot car is supposed to take a lane-changing action.

\section{Experiments}
In this section, we discuss the model calibration and validation. Some details of model implementation, calibration setup, and data are described, and the results are given.
\subsection{Data}
Our traffic flow data is collected by drones which have bird-view perspective. All vehicles' position, velocity and acceleration can be assumed accurate. And the locations of all scenarios covered lots of different cities in the US and in China. Different traffic conditions and scenarios, such as highway and downtown, are also included. Local maps with high definition are also built to get routing information.
\subsection{Implementation}
A high-definition map framework Lanelet2\cite{poggenhans2018lanelet2} is used to load map information and generate all possible routing path. A real-time route recommendation system is implemented to simulator the navigation system in each car, it collects traffic info and provides the drivers travel time of different routing path.
The platform is self-developed simulator, it can not only load real car's trajectory in real-time to calibrate our model but also generate robot cars to test the validity of our model. A renderer based on Unreal Engine 4 is used to visualize the result.
In validation part, we generate robot agents from random entrance of our map and each agent randomly choose a destination(exit in our map). The time gap of generation varies according to time arrival model \cite{may1990traffic}. These robot agents are equipped with different car-following model as we discussed previously.
\subsection{Calibration of parameters}
In the past studies of calibration, Maximum likelihood Method and RMSE are widely used to find the optimal parameters of the model\cite{knoop2014calibration}. In these methods, calibration is aimed to find the optimal parameters for minimizing the error. Here we use logistic regression to find out two set of parameters, one for aggressive drivers and another for conservative.

In our experiments, we firstly collected all cases drivers change to an asymmetric lane. Some of them back to their original lane, which means they change lane for overtaking(speed incentive). Others' behaviour were categorized as route change. Thus we can calibrate out the probability of back $P_b$ with $\alpha = 0.058$.

Then for each agent of our data, we can obtain their lane-changing interest as $$ w = A\mu_1 + B\mu_2 + C(\beta)\mu_3 + D\mu_4 $$
where A, B, D are constants which are calculated from data, only C is related to the parameter $\beta$. Thus we have only 5 parameters to calibrate. Additionally, a sigmoid function $ h(w) = \frac{1}{1+e^(-w)} $ is applied to approach the lane-changing probability. We use 1 for the vehicle changed their lane and 0 for the others. Thus we can form our logistic regression cost as:
$$Cost(h(w), y) = -y\log{h(w)} - (1 - y)log(h(w)) $$

Calibration consists of ﬁnding these values for the parameters, which minimize the logistic regression error J. 
$$J = \frac{ \sum{(Cost)}}{N}$$ where N is the total number of vehicles in our dataset. We use $\frac{2}{3}$ of our data to calibrate and the remaining data to test the validation. 
    
In fact, real drivers have distinguishing driving style as we discussed previously. Thus a set of fixed parameters is never enough to describe different people's behaviour. Thus we selected aggressive vehicles out, then calibrated them separately. The ratio between the aggressive drivers and conservative drivers is about 1 to 7.4. The result of calibration is showed as follows. Then we use these two set of weights in our simulator which displayed in the next subsection, the result shows our traffic flow generated are plausible for the macroscopic lane-changing probability.

\begin{table}[h]
\begin{center}
  \caption{Optimal parameters \label{tab:reg}}
    \begin{tabular}{|lcc|}
    \hline
    Parameters  &    Aggressive  & Conservative                     \\
    \hline\hline
    $ \alpha $   &     $ 0.058 $  &  $0.058$      \\
    
    $ \beta $    &     $ 1.1 $    &  $2.3$ \\
    
    $ \mu_1 $    &     $ 1.26 $   &  $0.44$  \\
    
    $ \mu_2 $    &     $ 0.58 $   &  $0.41$ \\     
     
    $ \mu_3 $    &     $ 0.03 $   &  $0.09$ \\
    
    $ \mu_4 $    &     $ 0.61 $   &  $1.72$ \\
    \hline
    \multicolumn{2}{l}{\scriptsize } \\
    \end{tabular}%
    \end{center}
\end{table}%

One of the great advantages of our model is the variable tuning parameters can produce totally different driving style in real world.
\subsection{Result}
we tested our model in two scenarios, highway and urban environment. The result shows in highway whose traffic density is low and the route change cost is relatively very high so that few driver tend to change their original route. In this case, our model has similar performance with other well-know lane-changing model, such as MOBIL \cite{kesting2007general}.However in urban environment, the result of our model shows the lane-changing rate in different traffic density fits well the observation of our dataset. \cite{knoop2016quantifying}. MOBIL doesn'n work well because it has to keep in right route.

The lane-changing rate is then given by:
$$r(\rho) = \frac{n}{\triangle{x}\triangle{t}}$$
Additionally, the number of lane-changing increase sharply while we augment the weight of route change incentive. This result is more in line with our expectations. Once the traffic is congested, people tend to change their original route for reducing the travel time. And in the future, more incentives are took into account, driving behaviour of a robot car could be more realistic and diverse. We would like to make the robot agents more intelligent that they can adjust their behaviour according traffic condition.

\begin{figure}[t]
\begin{center}
   \includegraphics[width=0.9\linewidth]{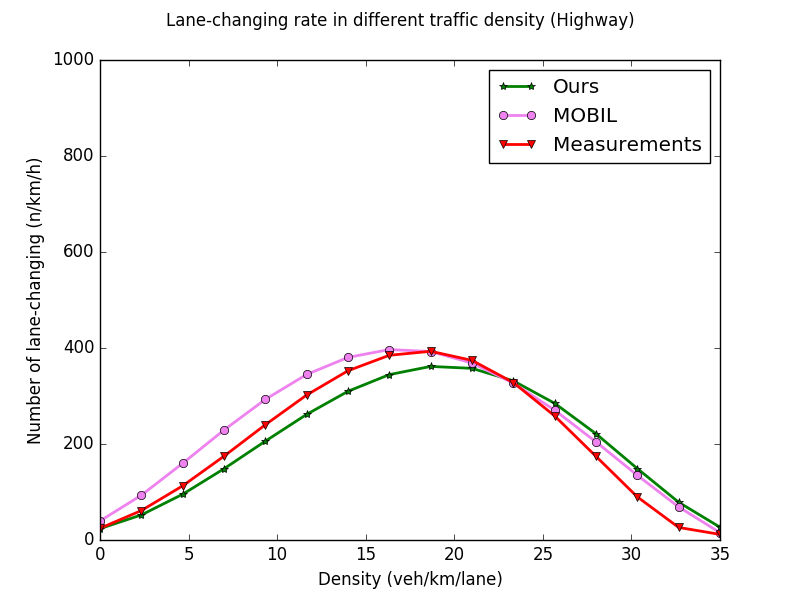}
\end{center}
   \caption{The experiment results demonstrates our model can produce similar lane-changing rates with MOBIL in highway. }
\label{fig:long}
\label{fig:onecol}
\end{figure}

\begin{figure}[t]
\begin{center}
   \includegraphics[width=0.9\linewidth]{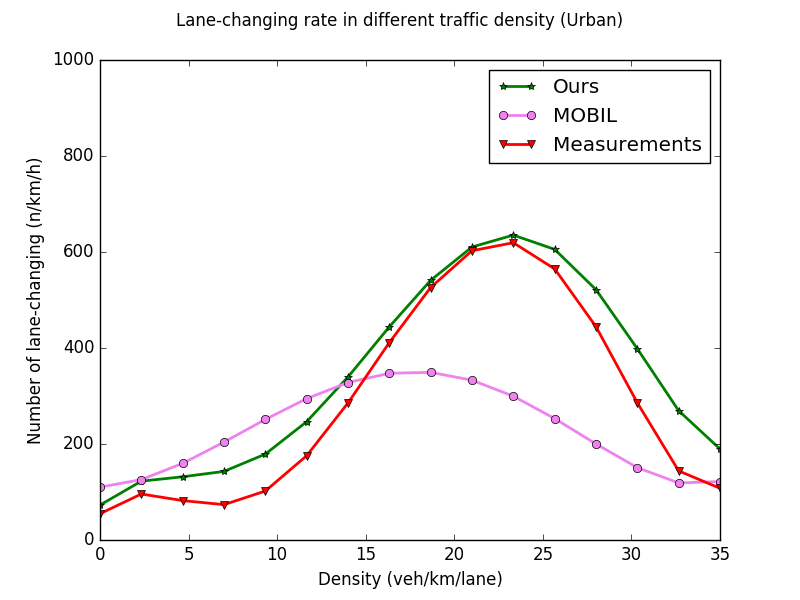}
\end{center}
   \caption{In urban environment, our model can produce more plausible lane-changing rates which are similar to the real data.}
\label{fig:long}
\label{fig:onecol}
\end{figure}

\section{Conclusion}
A methodology of building lane change model was proposed, the lane-changing decision lies on the drivers' own incentives. Within multiple-criteria, the combination of incentives can easily realized and convenient to mimic different chauffeur type by changing the strategy. The different driving styles are discussed and realised by tuning weights between different incentives. And different car-following models are integrated to simulate longitudinal motion. Each incentive based cost evaluation function are formalized and calibrated by real data. Validation shows the correspondence between driving style and lane-changing behaviour.

Our modbel can represent a more abundant and unexpected lane changing behavior with different preferences of incentives. And compared with artificial intelligence-based model, our model has more flexible controllability due to we can adjust the weights in multiple-criteria decision making. The model was calibrated and validated in both free-flow and congested urban traffic conditions. Furthermore, more other incentives can be took into account to create more various driving behaviours.







\bibliographystyle{IEEEtran}
\bibliography{IEEEabrv,IEEEexample}

\end{document}